\documentclass[pdflatex,sn-mathphys-num]{sn-jnl}

\usepackage{graphicx}%
\usepackage{multirow}%
\usepackage{amsmath,amssymb,amsfonts}%
\usepackage{amsthm}%
\usepackage{mathrsfs}%
\usepackage[title]{appendix}%
\usepackage{xcolor}%
\usepackage{textcomp}%
\usepackage{manyfoot}%
\usepackage{booktabs}%
\usepackage{algorithm}%
\usepackage{algorithmicx}%
\usepackage{algpseudocode}%
\usepackage{listings}%

\theoremstyle{thmstyleone}%
\theoremstyle{thmstyletwo}%

\theoremstyle{thmstylethree}%

\begin{document}

\title[Article Title]{AgriGov: A Structured Multilingual Dataset Curation for Indian Government Schemes for Farmers}

\author*[1]{\fnm{Mohsina} \sur{Bilal}}\email{mohsina\_m240456cs@nitc.ac.in}

\author[1]{\fnm{Gopakumar} \sur{G.}}\email{gopakumarg@nitc.ac.in}

\affil*[1]{\orgdiv{Computer Science and Engineering Department}, \orgname{National Institute of Technology Calicut}, \orgaddress{\city{Kozhikode}, \postcode{673601}, \state{Kerala}, \country{India}}}


\abstract{AgriGov is a curated, trilingual (English–Hindi–Marathi) dataset designed to address the scarcity of domain-grounded multilingual resources for agricultural policies and farmer welfare schemes. Initially, we collected and structured data from 50 government schemes sourced from trusted portals using automated scraping techniques, organizing it into predefined semantic fields (e.g., title, eligibility, application process, documents, exclusions). Translations were performed using a pipeline combining Google Translate API, MarianMT, and human post-editing, resulting in a domain-specific Hindi–Marathi dataset comprising \~2200 source segments. To enhance coverage, we augmented this dataset with sentences from the Samanantar corpus, leading to approximately 8,000 sentence-aligned Hindi–Marathi parallel pairs. The dataset now offers robust resources for fine-tuning machine translation models in this domain. AgriGov is designed for applications in domain-adaptive machine translation, question answering, information retrieval, and summarization systems. Its key contribution is a schema-driven, human-corrected multilingual alignment pipeline that ensures domain fidelity, provides provenance, and supports reproducible experiments, enabling retrieval-augmented applications for farmer-facing tools.}

\keywords{Multilingual NLP, Agricultural Policy, Machine Translation, Domain-Specific Datasets, Farmer Welfare} 



\maketitle
\section{Introduction}\label{sec1}

Government schemes are a cornerstone of India’s public welfare architecture, providing essential financial, technical, and institutional support to millions of farmers. However, the fragmented, unstructured, and language-barriered nature of scheme information, often available only in English and scattered across ministry portals or press releases, poses a significant barrier to effective access, especially for rural populations who primarily speak regional languages like Hindi and Marathi. This lack of accessibility hampers efforts to improve agricultural outcomes and economic resilience.

From an NLP perspective, this domain—characterized by formal, formulaic language and specialized terminology—remains underserved. While popular corpora for Indian languages (e.g., Samanantar, IndicTrans2, MuRIL) exist, they do not capture the unique lexical, syntactic, and semantic intricacies of government documents, such as eligibility clauses, procedural steps, and domain-specific entities. Consequently, off-the-shelf machine translation (MT) systems and information retrieval (IR) tools often fail to deliver reliable outputs for farmer-facing services, relying on fragile live-scraping pipelines that are difficult to reproduce and scale.

In response, this paper presents AgriGov, a reusable, trilingual corpus specifically curated for Indian government schemes in the domain of agriculture and farmer welfare. Unlike general-purpose corpora, AgriGov is tailored to capture the unique characteristics of agricultural policy language, providing structured, multilingual content in English, Hindi, and Marathi. The dataset is designed for domain-adaptive machine translation, question answering, information retrieval, and retrieval-augmented generation (RAG) systems, facilitating localized guidance for farmers across India.

Our curation pipeline integrates both automated and manual stages to ensure scale and quality. The process involves: (1) Scraping government scheme resources including myscheme.gov, Wikipedia, and the official government website for Agricultural Welfare in India; (2) Structuring data into semantic fields (e.g., title, eligibility, application process); (3) Converting records to JSON and segmenting into sentence-level units; (4) Translating drafts into Hindi using Google Translate/MarianMT, followed by human post-editing; (5) Expanding the corpus into Marathi via machine translation + human correction; (6) Masking placeholders (scheme names, identifiers) for domain fidelity; (7) Performing sentence-level segmentation and aligning Hindi–Marathi pairs; (8) Detecting and correcting misalignments through heuristics and manual review; (9) Restoring masked tokens to preserve context.

AgriGov is positioned as a seed dataset for building domain-specific NLP applications such as machine translation, question answering systems, and farmer-facing chatbots. The dataset not only supports reliable, reproducible development but also offers the flexibility to be extended with general linguistic data (via augmentation and integration with corpora like Samanantar) to improve coverage without sacrificing domain specificity. By providing provenance and structure, AgriGov enables the creation of scalable, localized solutions to enhance access to government welfare schemes for farmers.

\section{Related Works}\label{sec2}

Large-scale parallel corpora for Indic languages have played a central role in recent advances in multilingual NLP for the Indian subcontinent. Samanantar, currently the largest publicly available English–Indic bitext collection, aggregated tens of millions of sentence pairs by combining publicly released corpora with web-mined parallel sentences, demonstrating both the feasibility and power of large-scale web mining for Indic languages \cite{1}. These resources have enabled strong general-purpose multilingual models, but they are broadly domain-agnostic and thus do not capture the specialized lexicon, syntactic patterns, and document structures typical of government policy texts—such as scheme descriptions, eligibility clauses, and legalistic phrasing. Consequently, models trained primarily on such general corpora often underperform on policy- and procedure-oriented content that requires domain-specific terminology and pragmatic interpretation.

Recent work on pretrained representations for Indian languages (for example, MuRIL) and language-agnostic sentence embedding methods (for example, LaBSE-style approaches) has improved cross-lingual transfer and retrieval. MuRIL demonstrates that monolingual and multilingual pretraining on Indic text yields substantial gains for many downstream tasks compared to models trained only on non-Indic multilingual data \cite{2}. LaBSE-style and related sentence-embedding methods provide strong cross-lingual retrieval baselines and are commonly used as retrievers in RAG pipelines \cite{3}. However, these models are trained on general-domain corpora and therefore may not optimally capture policy semantics or the lexical idiosyncrasies of farmer-scheme documents—for example, subsidy-specific terms, eligibility formulations, or locally used administrative phrasings—motivating the creation of embeddings derived from domain-specific Hindi corpora to improve retrieval precision in policy-oriented RAG systems.

Multilingual NMT architectures and publicly released models (notably IndicTrans2) provide high-quality many-to-many translation capabilities for the scheduled Indic languages \cite{4},\cite{5}. Using such pretrained multilingual models as a base and fine-tuning them on in-domain parallel data is a well-established strategy for domain adaptation in machine translation: general pretraining supplies broad linguistic knowledge while domain-specific fine-tuning corrects terminology and style for targeted text types. Prior work on domain adaptation and mixed fine-tuning shows that mixing domain and out-of-domain corpora—rather than naively fine-tuning only on domain data or only on general data—often yields more robust systems that still generalize. For Marathi-Hindi translation in particular, leveraging an IndicTrans2 backbone and then fine-tuning on curated scheme-specific bitexts is therefore a principled approach to improve accuracy on policy texts while retaining general translation quality.

Back-translation (using monolingual target-language text to generate synthetic source sentences) is one of the most influential techniques for improving translation in low-resource settings: it effectively scales parallel data using monolingual resources and has been shown to yield consistent BLEU improvements \cite{6}. Complementary augmentation methods—such as simple text-level operations (synonym replacement, insertion, deletion, swapping) popularized by EDA—help increase lexical diversity and model robustness \cite{7}. For Indic languages, augmentation requires additional attention to script normalization and transliteration noise. Iterative and cross-directional back-translation (for example, Marathi→Hindi→Marathi loops) can further diversify synthetic data while preserving domain semantics. These augmentation strategies are particularly valuable for domain-specific corpora where parallel data is limited but monolingual domain text can be collected at scale.

Domain adaptation research has highlighted several effective recipes—sequential fine-tuning (general→domain), multi-task mixing, and mixed fine-tuning (training on a controlled blend of domain and general data). Empirical studies show that a controlled mixture can prevent catastrophic forgetting of general language patterns while ensuring the model learns domain-specific terminology. Applying this insight to the policy/citizen-scheme domain suggests a pragmatic mixing strategy: a higher proportion of domain-specific scheme text to teach policy vocabulary, complemented by general English–Indic bitexts (for example, Samanantar) to retain grammatical and fluency properties. This hybridization both stabilizes training and improves downstream transfer to real-world citizen-facing tasks.

Retrieval-augmented generation (RAG) frameworks, which combine a parametric language model with a non-parametric retriever over external text, are now standard for knowledge-intensive tasks because they reduce hallucination and provide provenance for responses. RAG performance depends critically on the retrieval component; domain-tuned retrieval (i.e., retrievers trained on domain-specific embeddings) typically outperforms generic retrieval when the target knowledge is specialized. For government scheme QA and chatbot assistants aimed at farmers, a RAG pipeline that retrieves from Hindi (or Marathi) domain-authenticated scheme texts—using dense domain embeddings extracted from the curated corpus—is likely to yield more accurate, relevant, and verifiable answers than a generic LLM operating alone. This motivates the creation of policy-specific embedding sets alongside the parallel corpus.

Government publications are often disseminated in heterogeneous formats (HTML pages, scanned PDFs, press releases). Large-scale corpus efforts demonstrate the value of combining web-crawled monolingual corpora, OCR for scanned documents, and robust alignment approaches to extract high-quality parallel sentences from noisy sources. Tools such as Tesseract remain practical options for OCR processing of scanned PDFs, and careful post-processing (deskewing, normalization, filtering) is essential to ensure textual fidelity. These techniques underpin web-scale corpus construction and are directly applicable to harvesting scheme documents across multiple government portals and state-level sites.

Benchmarking translation and retrieval requires a combination of lexical and semantic metrics: BLEU and ChrF provide surface-level comparators, while neural metrics such as COMET correlate better with human judgment for modern systems \cite{8}. For low-resource evaluation, curated benchmark sets such as FLORES-101/FLORES-200 provide standardized testbeds for many languages and can serve as reference points when reporting multilingual performance. Finally, transparent dataset documentation (e.g., Datasheets for Datasets) is now considered best practice, facilitating reproducibility and ethical reuse by listing provenance, licensing, intended uses, and limitations—all crucial for government-sourced corpora.

In summary, prior work provides robust tools and methods—large general-purpose bitexts, pretrained multilingual encoders, NMT backbones, augmentation recipes, and retrieval frameworks—but there remains a clear gap: a publicly available, metadata-rich, domain-specific parallel corpus for Indian government farmer schemes that is trilingual (English–Hindi–Marathi) and designed for MT fine-tuning, domain-aware embeddings, and RAG-based citizen-facing systems. This gap motivates our AgriBhasha/AgriGov-Corpus: it combines domain curation, structured metadata, controlled domain–general mixing, tailored augmentation, and downstream embedding/retrieval benchmarks to enable reproducible research and practical deployments for policy-aware NLP in Indian languages.

\section{Methodology}\label{sec3}

\subsection{Data Description}\label{subsec3}

The AgriGov dataset is a structured trilingual corpus (English–Hindi–Marathi) developed to advance multilingual Natural Language Processing (NLP) research in the domains of agriculture, farmer welfare, and public policy communication. It captures official descriptions of government schemes targeted at farmers in India, encompassing sections such as scheme overviews, eligibility criteria, procedural guidelines, and benefit explanations. Emphasizing linguistic accessibility and inclusivity, AgriGov provides sentence-level aligned and semantically consistent text across all three languages, enabling comparative linguistic analysis and machine translation benchmarking in low-resource Indic contexts. Designed with reusability and standardization in mind, the dataset offers a well-documented schema detailing its provenance, structure, and linguistic coverage, supported by comprehensive descriptive statistics. By bridging the gap between government communication and computational linguistic research, AgriGov serves as a foundational resource for developing and evaluating multilingual models, alignment techniques, and domain-specific language understanding systems in the agricultural policy domain.

\begin{table}[h]
\caption{AgriGov Dataset Overview}\label{tab1}%
\begin{tabular}{@{}ll@{}}
\toprule
\textbf{Attribute} & \textbf{Description} \\
\midrule
Languages & English (EN), Hindi (HI), Marathi (MR) \\
Domain & Agricultural and farmer welfare schemes \\
Sources & Verified Indian government portals and ministry sites \\
Data Scale & 50 unique schemes, $\sim$2,100 raw segments, $\sim$8,000 sentence pairs \\
Use Cases & Machine Translation (HI$\leftrightarrow$MR), Multilingual QA, \\
          & Information Retrieval, RAG-based assistants \\
\botrule
\end{tabular}
\end{table}
This dataset is intended as a reusable benchmark for low-resource machine translation, domain adaptation, and multilingual model evaluation for Indian languages.

\begin{table}[h]
\caption{End-to-end pipeline stages for AgriGov dataset curation}
\label{tab:pipeline}
\centering
\begin{tabular}{@{}p{3cm}p{5cm}p{5.5cm}@{}}
\toprule
\textbf{Stage} & \textbf{Purpose} & \textbf{Output} \\
\midrule
Source Discovery & Identify authoritative scheme pages from official portals (myscheme.gov.in, Ministry of Agriculture \& Farmers Welfare) while preserving provenance metadata & Curated list of scheme URLs \\
Scraping & Extract structured text content from heterogeneous websites using a hybrid scraping strategy (dynamic + static sources) with ethical rate limits & Raw textual records stored for processing \\
Schema Mapping & Convert unstructured narrative text into standardized semantic fields aligned with farmer information needs & Schema-structured dataset (Title, Eligibility, etc.) \\
Segmentation & Produce sentence or clause-level units suitable for downstream multilingual alignment & Segmented records with scheme-level IDs \\
Masking & Temporarily replace domain-specific entities (scheme names, numeric identifiers) to protect them during translation & Masked multilingual JSONL with mapping for restoration \\
MT Drafts & Generate initial Hindi and Marathi translations via pivot-based machine translation pipeline with human monitoring & Draft English–Hindi–Marathi triplets \\
Human Post-editing & Refine translations for accuracy, terminology, and contextual correctness in policy language & Corrected bilingual and trilingual parallel data \\
Alignment \& QA & Perform Hindi–Marathi alignment with heuristic checks and manual review to ensure 1:1 segment correspondence & Aligned and validated sentence pairs with diagnostic reports \\
Augmentation & Expand corpus diversity through controlled back-translation or paraphrasing methods for domain adaptation & Synthetic bitexts and mixed corpora for MT fine-tuning \\
Release & Package final dataset with attribution, provenance, and licensing for public research & Public aligned dataset with documentation and changelog \\
\botrule
\end{tabular}
\end{table}

\subsection{Data Curation Pipeline}\label{subsec3}

\begin{figure}[h]
\centering
\includegraphics[width=0.9\textwidth]{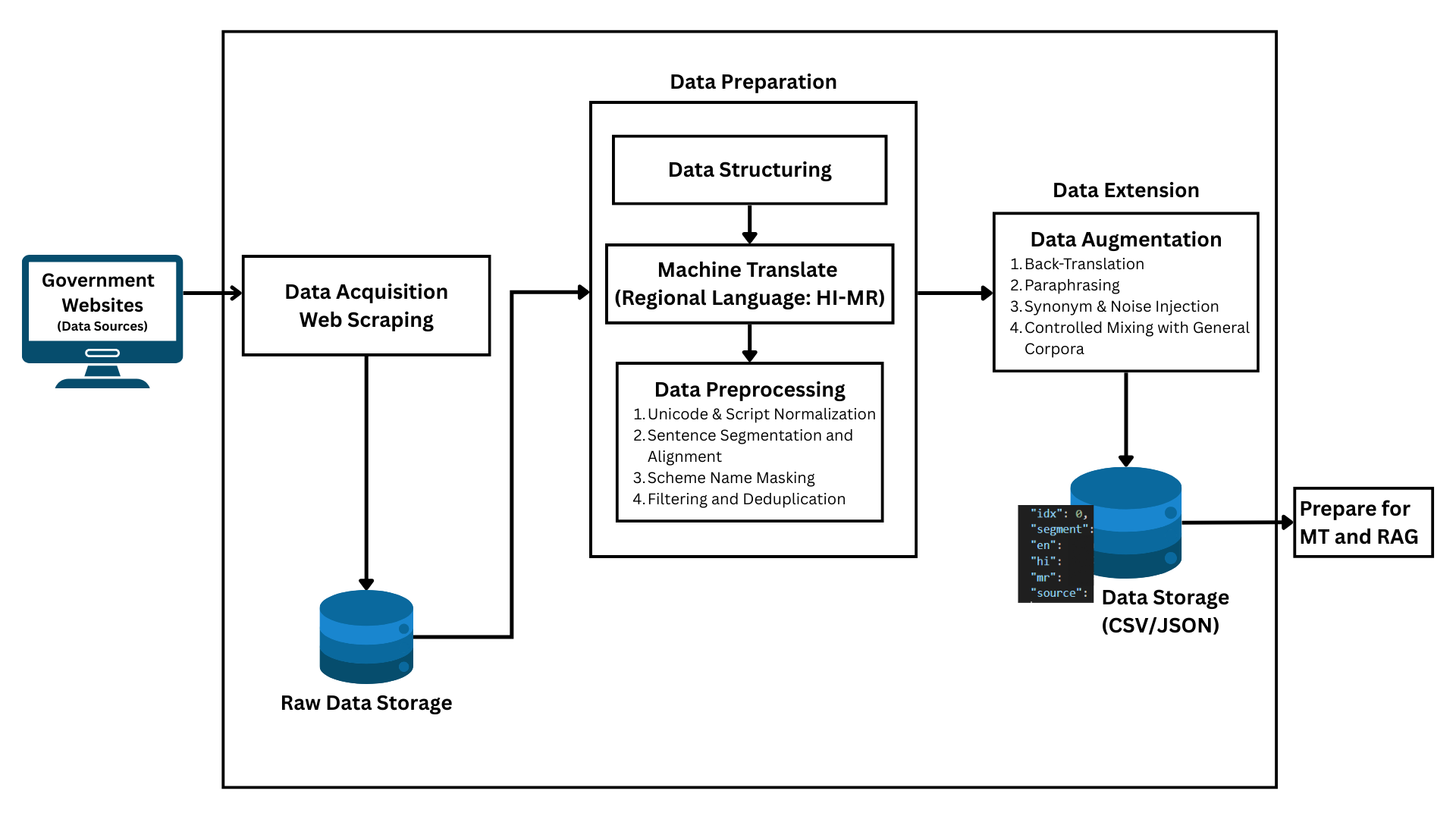}
\centering
\caption{End-to-end data curation pipeline}\label{fig:pipeline}
\end{figure}

AgriGov is constructed through a streamlined curation workflow that combines automated extraction with targeted human validation to ensure domain accuracy. The process includes scraping English scheme content from authoritative portals, structuring it into a standardized schema, segmenting records into sentence-level units, translating and post-editing bilingual outputs, masking and restoring sensitive tokens, and performing Hindi–Marathi alignment with quality checks. The pipeline is fully versioned and provenance-aware, retaining intermediate artifacts to ensure reproducibility and controlled augmentation.

Table~\ref{tab:pipeline} summarizes the core stages of the AgriGov data curation pipeline presented in this section.

\subsubsection{Data Acquisition}\label{subsubsec3}

The dataset begins with the collection of English-language scheme descriptions that directly target agriculture and farmer welfare audiences in India. To ensure reliability and factual correctness, only official and publicly accessible government-backed resources were used. The core content was sourced from myscheme.gov.in, which serves as the primary national portal aggregating centrally sponsored and central sector schemes. This was complemented with information from the Ministry of Agriculture \& Farmers Welfare website, which provides more detailed program guidelines, benefit structures, and implementation processes.\cite{9}, \cite{10}

Additionally, Wikipedia was selectively referenced only when it cited official government sources, and such content was used strictly for contextual introduction rather than as primary policy text. Across these sources, the acquisition process focused on extracting high-value fields including scheme overviews, eligibility criteria, application procedures, required documents, and benefit details — ensuring comprehensive coverage of policy information most relevant to farmer support and entitlements.\cite{11}

\begin{table}[h]
\caption{Semantic schema for structuring scheme content}\label{tab:schema}
\centering
\begin{tabular}{@{}ll@{}}
\toprule
\textbf{Field} & \textbf{Description} \\
\midrule
Title & Scheme name \\
Eligibility & Who can apply \\
Application\_Process & Steps and procedure to apply \\
Documents\_Required & Administrative documents needed \\
Exclusions & Ineligible cases or disqualifying conditions \\
Policy\_Details & Benefits, financial support, and key provisions \\
Source & Provenance URL of the official page \\
\botrule
\end{tabular}
\end{table}

\subsubsection{Data Processing and Structuring}\label{subsubsec3}

Government scheme content is typically presented as unstructured narratives or semi-formal web pages. To enable machine readability and downstream NLP workflows, the collected text was standardized into a domain-specific schema representing the core informational needs of farmers. Each scheme was divided into clearly defined fields including: \textit{Title}, \textit{Eligibility}, \textit{Application Process}, \textit{Documents Required}, \textit{Exclusions}, \textit{Policy Details}, and \textit{Source} as per table \ref{tab:schema}. This structure supports semantic consistency and facilitates fine-grained retrieval and supervision for tasks such as question answering and machine translation.

To improve coherence, normalization operations were applied across languages, including removal of boilerplate website elements, unifying punctuation and whitespace conventions, and standardizing numeric expressions and government terminology. Following normalization, each field was segmented into sentence-level units while preserving the logical grouping of legal and procedural clauses. This ensured that each resulting segment represents a complete actionable meaning, ready for multilingual expansion and alignment.

\subsubsection{Multilingual Expansion, Alignment and Quality Assurance}\label{subsubsec3}

Multilingual transformation began with Hindi as the base regional language due to its widespread official use across India. A two-stage machine translation strategy was used, first generating Hindi translations from English segments, then producing Marathi translations from the Hindi text to leverage linguistic similarity. Human post-editing was performed selectively to correct agricultural terminology, government policy vocabulary, and contextual ambiguities.

To maintain terminological precision, named entities such as scheme titles, government identifiers, financial amounts, and URLs were temporarily replaced with structured placeholders prior to translation. These placeholders were reinstated after multilingual alignment to ensure domain fidelity.

Sentence-level Hindi–Marathi alignment was conducted using heuristic checks such as segment length ratios and token-type verification to identify mistranslations, omissions, or non-linguistic noise. Flagged pairs were reviewed manually, with emphasis on eligibility conditions and rule-driven clauses. Final quality validation included reconciliation of placeholders, numeric equivalence checks, and contextual consistency across languages. This phase produced a high-precision trilingual dataset with preserved provenance for reproducible experimentation.

\section{Results}\label{sec4}

To demonstrate the quality and practical usefulness of the AgriGov dataset, we present qualitative samples from different stages of the curation pipeline along with baseline performance insights. Together, these results validate that AgriGov is a reliable, domain-accurate resource capable of supporting multilingual modeling in agricultural policy contexts.

\subsection{Data Statistics}\label{subsec:4}

The AgriGov dataset contains English–Hindi–Marathi parallel content for 50 centrally relevant agricultural welfare schemes, resulting in over 2,100 segmented records and approximately 8,000 Hindi–Marathi aligned sentence pairs. Each scheme is represented across seven standardized semantic fields, ensuring balanced distribution and comprehensive linguistic coverage.

\begin{figure}[h]
\centering
\includegraphics[width=0.95\textwidth]{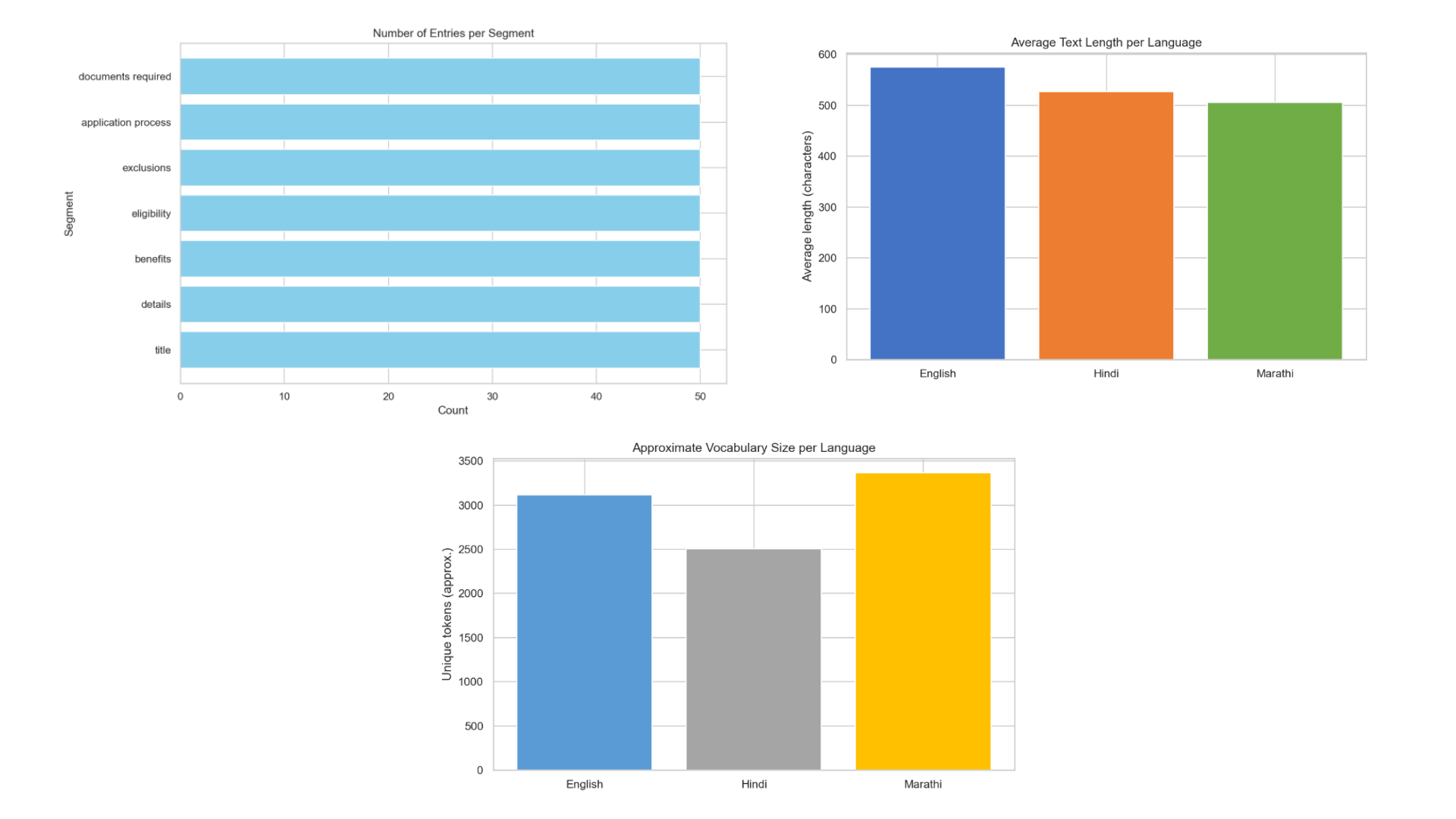}
\caption{Dataset statistics across semantic fields and languages: (a) segment distribution, (b) average text length, (c) vocabulary diversity.}\label{fig:stats}
\end{figure}

To characterize the corpus, we summarize segment counts, text length distributions, and vocabulary diversity across all three languages. As shown in Figure~\ref{fig:stats}, English entries tend to be slightly longer on average, whereas Marathi exhibits greater lexical variation due to morphological richness. Across languages, distribution curves follow a long-tailed pattern, with most policy statements under 500 characters but some procedural descriptions extending beyond 2,000 characters. Data cleaning eliminated artifacts such as HTML remnants and non-linguistic noise, preserving a high-quality text source traceable to official portals. Overall, the dataset provides a reliable foundation for multilingual modeling, alignment experiments, and domain-adapted translation research.

\subsection{Dataset Output Samples}\label{subsec:4}

To demonstrate the final structure and semantic quality of the corpus English–Hindi–Marathi aligned segments from the AgriGov dataset are analyzed. These samples highlight preservation of key policy semantics such as eligibility rules, numerical constraints, and administrative terminology aspects where generic MT systems often fail.

\begin{figure}[h]
\centering
\includegraphics[width=0.85\textwidth]{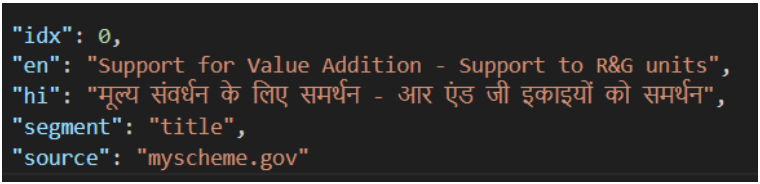}
\caption{Example of structured English-Hindi JSON record after data extraction, formatting and translation to pivot lang.}
\label{fig:raw_json}
\end{figure}

\begin{figure}[h]
\centering
\includegraphics[width=0.85\textwidth]{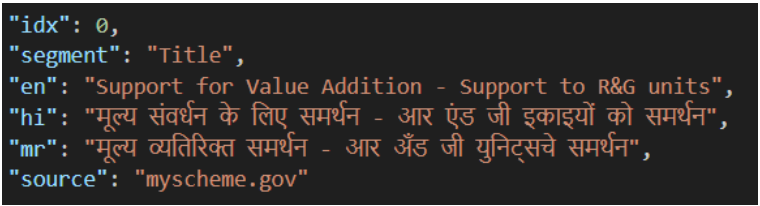}
\caption{Example trilingual aligned record (EN--HI--MR) after multilingual expansion and post-editing.}
\label{fig:mr_expansion}
\end{figure}

Figure~\ref{fig:raw_json} illustrates a sample JSON record immediately after initial processing, showing structured English fields extracted from official sources. Following multilingual expansion and human post-editing, Figure~\ref{fig:mr_expansion} displays the corresponding trilingual aligned version of the same segment, demonstrating consistency in terminology and meaning across languages.

These qualitative samples confirm that AgriGov maintains domain fidelity and supports downstream tasks such as machine translation, cross-lingual retrieval, and RAG-based policy information assistants.

\subsection{Alignment Quality and Language Integrity}

The Hindi–Marathi alignment process relied on automatic heuristics to detect translation inconsistencies, structural divergences, and malformed outputs, ensuring that only a small fraction of segments—mainly those involving numeric constraints or procedural clauses—required manual inspection. A focused human audit confirmed that the final alignment accuracy reached approximately 90\%, demonstrating the reliability of the heuristic-driven approach. Additionally, the use of placeholder masking effectively prevented the mistranslation of named entities and financial amounts, preserving essential policy terminology across languages. Importantly, all seven semantic fields maintained balanced representation in the aligned corpus, reducing bias toward narrative components and ensuring comprehensive coverage of eligibility, benefits, documents, and procedural information.

Table~\ref{tab:alignment_summary} showcases a diagnostic excerpt from the validation stage, including summary statistics and representative flagged cases requiring correction.

\begin{table}[h]
\centering
\caption{Hindi--Marathi Alignment Quality Summary}
\label{tab:alignment_summary}
\begin{tabular}{l r}
\toprule
\textbf{Metric} & \textbf{Value} \\
\midrule
Total sentences checked & 1901 \\
Mean length ratio (HI/MR) & 1.74 $\pm$ 4.44 \\
Potential misalignments & 252 (13.3\%) \\
Accurately aligned sentences & 1649 (86.7\%) \\
\bottomrule
\end{tabular}
\end{table}

These results validate that the multilingual expansion and alignment strategies used in AgriGov are effective in preserving structural and semantic correctness while minimizing translation noise.

\section{Discussion}\label{sec5}

The development of AgriGov underscores the need for domain-specific multilingual datasets within public-sector NLP, particularly for agriculture and farmer welfare in India. Existing Indic resources such as general-purpose parallel corpora often fail to represent the procedural and policy-specific terminology farmers interact with when accessing governmental support. In contrast, AgriGov focuses exclusively on structured agricultural welfare communication aligned across English, Hindi, and Marathi—three crucial languages for rural outreach.

A core novelty of AgriGov lies in its schema-driven representation, which segments scheme content into semantically meaningful fields including eligibility, application process, documents required, exclusions, and benefit details. This supports tasks beyond translation, such as answer grounding, structured retrieval, and rule-aware question answering. The decision to align the dataset with existing Indian-language bitext standards further enhances interoperability within multilingual model training pipelines.

Ethical sourcing and reproducibility are emphasized throughout the curation process. All content originates from legally accessible Indian government portals, maintaining data transparency and public trust. Human post-editing reinforces translation correctness for legal and numeric phrasing, mitigating common automated translation errors. The alignment pipeline reports and corrects segment-level divergences, contributing insight toward improving multilingual dataset quality in low-resource, semi-structured policy domains.

Finally, by providing a static, version-controlled release, AgriGov avoids the instability of dynamic web-scraping pipelines and supports fair benchmarking for future research in domain-grounded NLP.

\subsection{Comparison with Existing Datasets}\label{subsec5}

Table~\ref{tab:dataset_comparison} compares AgriGov with major existing Indic datasets. Whereas large-scale corpora such as Samanantar provide broad linguistic coverage, they lack field-level structure and domain specificity needed for accurate government policy interpretation. Other agricultural datasets do not provide parallel content, limiting multilingual system development.

\begin{table}[h]
\caption{Comparison of AgriGov with existing publicly available Indic datasets}
\label{tab:dataset_comparison}
\centering
\begin{tabular}{@{}p{2.5cm}p{2.5cm}p{2cm}p{2cm}p{3cm}@{}}
\toprule
\textbf{Dataset} & \textbf{Languages} & \textbf{Domain} & \textbf{Structure} & \textbf{Parallel Utility} \\
\midrule
Samanantar & EN–Indic & General text & Sentence-level & Yes (but not policy-specific) \\
Wikipedia-based corpora & EN–Indic & Open-domain & Sentence-level & Yes (generalized content) \\
Govt. scheme websites & EN or Indic & Policy text & Narrative & No parallel alignment \\
Agri-news datasets & HI/MR & Agriculture & Topic-driven & No parallel content \\
\textbf{AgriGov (ours)} & EN–HI–MR & Farmer welfare schemes & Field-structured schema & \textbf{Yes: 1:1 aligned} \\
\botrule
\end{tabular}
\end{table}

AgriGov introduces several innovations that distinguish it from existing resources. It is the first trilingual dataset dedicated specifically to Indian agricultural welfare schemes, structured through a schema-driven segmentation process that mirrors real farmer information needs. The dataset preserves policy-specific terminology—including eligibility conditions, subsidy rules, and administrative clauses—making it particularly valuable for legal and procedural contexts. It is designed to support both machine translation workflows and policy-aware knowledge access systems, with version-controlled updates and human-corrected entries ensuring accuracy, especially for numerical and regulatory content. Additionally, its structured and aligned format allows seamless integration into retrieval-augmented generation (RAG) pipelines, enabling high-quality, grounded responses in multilingual settings.

Overall, AgriGov establishes a high-quality, domain-rich, and practically deployable foundation for multilingual public information systems supporting Indian farmers.

\subsubsection{Applications and Future Work}\label{subsubsec5}

Building on the curated multilingual dataset, several directions are planned to expand the research and real-world deployment potential of AgriGov in agricultural communication systems. These efforts target improvements in translation quality, domain-aware retrieval, and integration into interactive language technologies for rural communities.

\textbf{Domain-Adaptive Machine Translation:} A key next step involves training baseline Hindi–Marathi machine translation models using AgriGov. Transformer-based architectures such as MarianMT and mBART \cite{12}, \cite{13} will be fine-tuned on the aligned corpus to capture policy-specific linguistic structure and terminology. This will demonstrate measurable gains over general MT systems trained exclusively on broad-domain corpora and establish benchmarks tailored to government scheme discourse.

\textbf{Embedding-Based Information Retrieval:} Dense semantic representations will be created for Hindi and Marathi segments using retrieval-focused encoders such as Sentence-BERT\cite{14} and LaBSE. These embeddings will power similarity-based lookups, allowing natural-language queries about eligibility, benefits, or procedures to retrieve the most relevant scheme passages. This lays the groundwork for knowledge-grounded retrieval systems that enhance public access to welfare information.

\textbf{Retrieval-Augmented Conversational AI:} Longer-term development includes integrating AgriGov into retrieval-augmented generation (RAG) frameworks built on Hindi-capable LLMs. Retrieved scheme snippets will supply factual grounding for chatbot responses, enabling reliable and verifiable answers to farmer queries. This approach supports trustworthy conversational agents aligned with government policy and rural language accessibility.

These future extensions will not only broaden AgriGov’s research impact but also accelerate the creation of practical, citizen-centric tools capable of improving informed participation in welfare programs across India.

\section{Conclusion}\label{sec6}

This work introduced AgriGov, a structured trilingual dataset focused on Indian government schemes for farmers, addressing critical gaps in multilingual agricultural communication resources. AgriGov compiles authoritative content from trusted government portals and represents it in a schema-driven format aligned across English, Hindi, and Marathi. Through a reproducible curation pipeline—encompassing targeted web acquisition, semantic field structuring, controlled machine translation with human post-editing, placeholder-based domain fidelity preservation, and rigorous sentence-level alignment—the dataset delivers over 5,000 high-quality parallel records tailored to downstream NLP applications.

AgriGov provides a foundational resource for improving domain-specific performance in machine translation, retrieval, summarization, and conversational AI. Planned expansions include data augmentation strategies (e.g., back-translation, paraphrasing), enriched alignment validation, and integration within retrieval-augmented generation (RAG) frameworks to support interactive policy assistance in regional languages. By prioritizing linguistic accessibility, transparency, and public-sector relevance, AgriGov contributes to the broader mission of democratizing AI for social welfare. It enables researchers and developers to build reliable, factual, and inclusive tools that enhance communication between government institutions and rural farming communities across India.

\backmatter

\bmhead{Acknowledgements}

The authors gratefully acknowledge the IndiaAI Initiative of the Ministry of Electronics and Information Technology, Government of India, for fellowship support.  They thank the National Institute of Technology Calicut, Department of Computer Science and Engineering for providing research facilities and administrative assistance. The authors have reviewed and edited the output and take full responsibility for the content of this publication.

\section*{Declarations}

\begin{itemize}
    \item \textbf{Funding:} This work was supported by the fellowship acquired from the IndiaAI initiative of the MIETY, Government of India.
    
    \item \textbf{Conflict of interest/Competing interests:} The authors declare that they have no conflict of interest.
    
    \item \textbf{Ethics approval and consent to participate:} This study does not involve human participants or animal subjects, so ethics approval is not applicable. If your research involved human or animal subjects, you would need to state the approval and consent process here.
    
    \item \textbf{Consent for publication:} The authors provide consent for publication of this manuscript.
    
    \item \textbf{Data availability:} The dataset used in this research is available in the public GitHub repository: https://github.com/mohsina-bilal/Farmer-Chatbot. The dataset includes all the relevant data required for replication of this study and future research.
    
    \item \textbf{Materials availability:} The materials used in this research (e.g., forms, documentation) are available upon request or can be accessed from the same GitHub repository mentioned above.
    
    \item \textbf{Code availability:} The code used for data processing, alignment, and machine translation is available upon request. Please contact the corresponding author for access.
    
    \item \textbf{Author contributions:} 
    \begin{itemize}
        \item Mohsina Bilal was responsible for the design and implementation of the data curation pipeline, including dataset acquisition, curation, analysis, and manuscript preparation.
        \item Gopakumar G provided governance oversight and contributed to the analysis and critical review of the manuscript.
    \end{itemize}
\end{itemize}


\bibliography{sn-bibliography}

\end{document}